\documentclass{article}

\usepackage{PRIMEarxiv}

\usepackage[utf8]{inputenc} 
\usepackage[T1]{fontenc}    
\usepackage{hyperref}       
\usepackage{url}            
\usepackage{booktabs}       
\usepackage{amsfonts}       
\usepackage{nicefrac}       
\usepackage{microtype}      
\usepackage{lipsum}
\usepackage{fancyhdr}       
\usepackage{graphicx}       
\graphicspath{{media/}}     
\usepackage{multirow}
\usepackage{caption}
\usepackage{CJK}
\title{ANGO: A Next-Level Evaluation Benchmark For Generation-Oriented Language Models In Chinese Domain} 

\author{
  Wang, Bingchao \\
  \texttt{wangbingchaofun@gmail.com} \\
}

\begin{document}
\maketitle

\begin{abstract}
Recently, various Large Language Models (LLMs) evaluation datasets have emerged, but most of them have issues with distorted rankings and difficulty in model capabilities analysis.
Addressing these concerns, this paper introduces ANGO, a Chinese multi-choice question evaluation benchmark. 
ANGO proposes \textit{Keypoint} categorization standard for the first time, each question in ANGO can correspond to multiple keypoints, effectively enhancing interpretability of evaluation results.
Base on performance of real humans, we build a quantifiable question difficulty standard and divide ANGO questions into 9 difficulty levels, which  provide more precise guidance for model training.
To minimize data leakage impact and fully leverage ANGO's innovative features, we have engineered exclusive sampling strategies and a new evaluation framework that support swift testset iteration.
Our experiments demonstrate that ANGO poses a stronger challenge to models and reveals more details in evaluation result compared to existing benchmarks.
\end{abstract}

\section{Introduction}
The field of Natural Language Processing (NLP) has witnessed significant advancements in recent years, largely attributed to the development of the LLM which possesses a vast number of parameters and trains on extensive corpus. A pivotal breakthrough is the introduction of the Transformer architecture, which incorporates the Attention mechanism \cite{vaswani2023attention}. This architecture facilitates the modeling of long-range dependencies in textual data and enables parallel training. Subsequently, pioneering models such as BERT \cite{devlin2019bert} emerged, setting new standards for self-supervised pretraining. More recently, decoder-only models like GPT \cite{brown2020language} have gained widespread adoption as the foundation for constructing LLMs.
\begin{figure}[ht]
\begin{center}
\centerline{\includegraphics[width=\columnwidth]{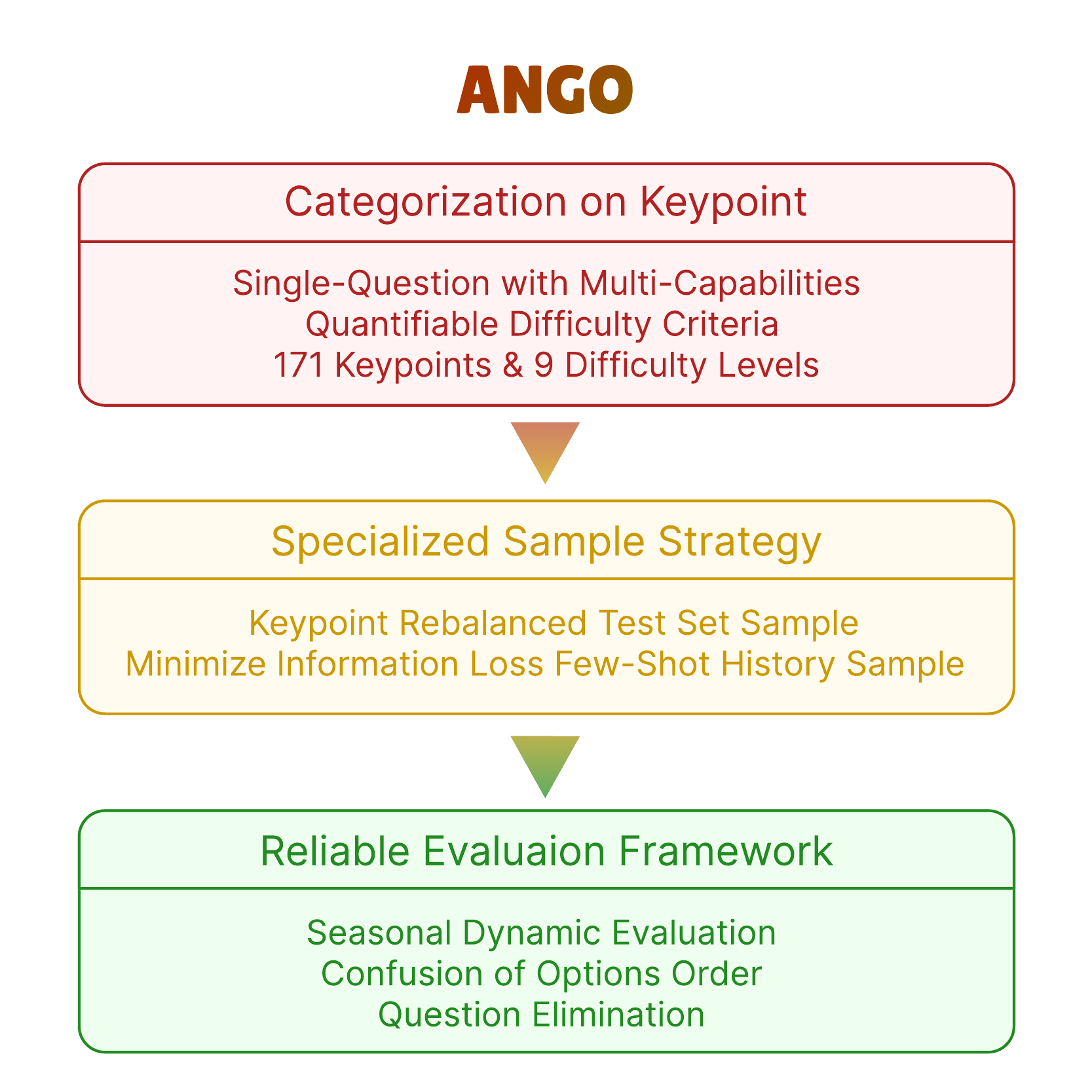}}
\caption{ANGO: A Next-level evaluation benchmark for Generation-Oriented language models}
\label{Introduction}
\end{center}
\end{figure}

As LLMs have progressed, the benchmarks used to evaluate them have also evolved. Initial works primarily focused on assessing Natural Language Understanding (NLU) tasks, exemplified by benchmarks like GLUE \cite{wang2019glue} and its multi-task variant superGLUE \cite{sarlin2020superglue}. However, as research has delved deeper into the linguistic capabilities of LLMs, specialized benchmarks with multiple-choice question formats have emerged to evaluate specific abilities such as reasoning and inference. Notable examples include MMLU \cite{hendrycks2021measuring}, TruthfulQA \cite{lin2022truthfulqa}, Big-Bench \cite{srivastava2023beyond} in English, and AGI-Eval \cite{zhong2023agieval}, C-Eval \cite{huang2023ceval}, CMMLU \cite{li2023cmmlu} for the Chinese domain, which are now widely adopted and prominently featured on leaderboards.

\paragraph{Challenges:} Existing multi-choice question benchmarks primarily rely on collecting exam questions to construct datasets and establish categories and difficulty levels based on the \textit{Subject} covered in these exams. However, this simplistic categorization suffers from the following general problems:

\begin{itemize}
\item \textbf{Single-Question Single-Subject}: Subject-based categories disregard that questions often require proficiency in multiple subjects and overlooking the existence of further subdivisions within each subject. 
\item \textbf{Immeasurable Difficulty}: Difficulty levels typically determined by the educational grade associated with the subject, despite significant variations in difficulty for different questions in same subject.
\item \textbf{Hard To Update}: Benchmark data leaks always result in serious distortion\cite{zhou2023dont}. Most of them are challenging to conduct periodic updates due to potential sampling biases.
\end{itemize}

Because of the aforementioned issues, traditional benchmarks struggle to provide reliable measurements for model's ability to handle a diverse range of disciplines and fail to accurately assess the performance for complex questions. 

\paragraph{Contributions:} We introduce ANGO as a novel Chinese benchmark for evaluation with the format of multiple-choice questions. ANGO apply \textit{Keypoint} instead of \textit{Subject} as the criteria to address previous challenges by following improvements:
\begin{itemize}
\item \textbf{Single-Question Multi-Keypoints}: Question in ANGO can correspond to 1-6 different keypoints, and include 988 keypoint combinations. The number of keypoints reach up to 171 distinct keypoints.
\item \textbf{Quantifiable Difficulty}: For difficulty assessment, we introduce a quantifiable function that incorporates real human performance, and divided questions in ANGO into 9 grades.
\item \textbf{Easy To Update}: We design specific strategies for sampling, which meet the balanced keypoints distribution and ensure minimal information loss when reducing the number of few-shot examples.
\end{itemize}

With the innovative features of ANGO, we build an evaluation framework that can provide more comprehensive and accurate multi-level, multi-perspective performance results for participating models. Furthermore, this framework allows dynamic updates of the testset to ensure the objectivity and authenticity.

\section{Data}
\subsection{Data Collection}
\paragraph{Data Source}
Our research data is exclusively sourced from the Administrative Proficiency Test (AAT) used in the Chinese civil service examination, which encompasses 34 provinces in China, including both official and mock exams conducted between 2008 and 2023.

The AAT is a comprehensive assessment tool comprised entirely of multiple-choice questions, designed to evaluate the abilities and skills required for practical administrative work. The language model exhibits remarkable performance in generating and comprehending text, and the AAT provides specific and intricate contexts that simulate communication and decision-making processes in the real world. By utilizing the language model to answer questions from the AAT, we can assess its understanding of complex problems, its ability for judgment and reasoning, as well as the accuracy and fluency of its language expression. Furthermore, the AAT covers a wide range of question domains and scenarios, this diversity contributes to evaluating the language model's language processing capabilities in different domains. 
\begin{figure}[ht]
\begin{center}
\centerline{\includegraphics[width=\columnwidth]{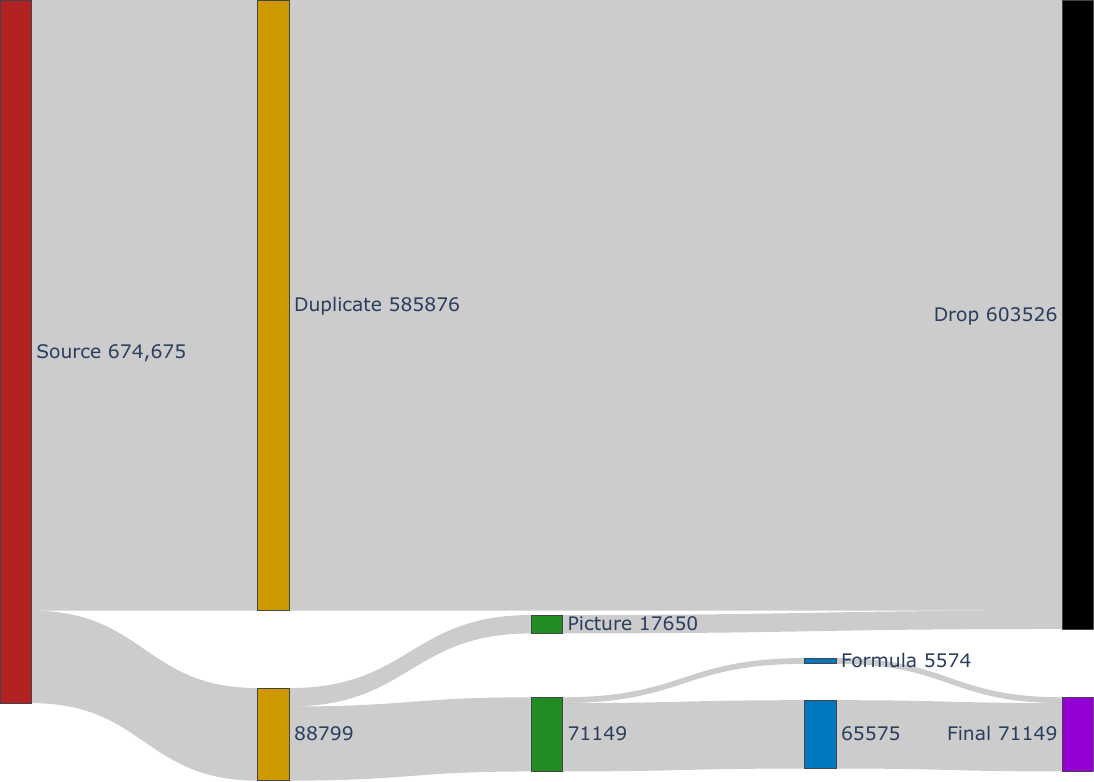}}
\caption{The Data Preprocess Flow: From 674,675 source data to 71,149 clean data}
\label{Data Process}
\end{center}
\end{figure}

\paragraph{Data Preprocessing}
After collecting without any manipulation, we obtained a total of 674,675 records. In order to enhance the quality of our data, we employed a series of simple but efficient preprocess functions, as shown in Figure \ref{Data Process}: (1) Mock exams often include previous existed exam questions, which lead to numerous duplicates. We remove duplicates based on the content of records; (2) Since our primary focus is on NLP instead of Multi-Modal, we removed all records containing pictures in both the questions and options; (3) By manual recognition aided by OCR model, we extracted 5,574 unique formulas from a pool of 34,062 \textbf{\LaTeX} formula images and convert them to text.

\subsection{Data Distribution}
After completing Data Collection and Data Preprocessing, we divided the records into different keypoints and difficulty levels based on the question content and the domains involved in them.
\subsubsection{Keypoint Tree}
\begin{figure}[ht]
\centerline{\includegraphics[width=\columnwidth]{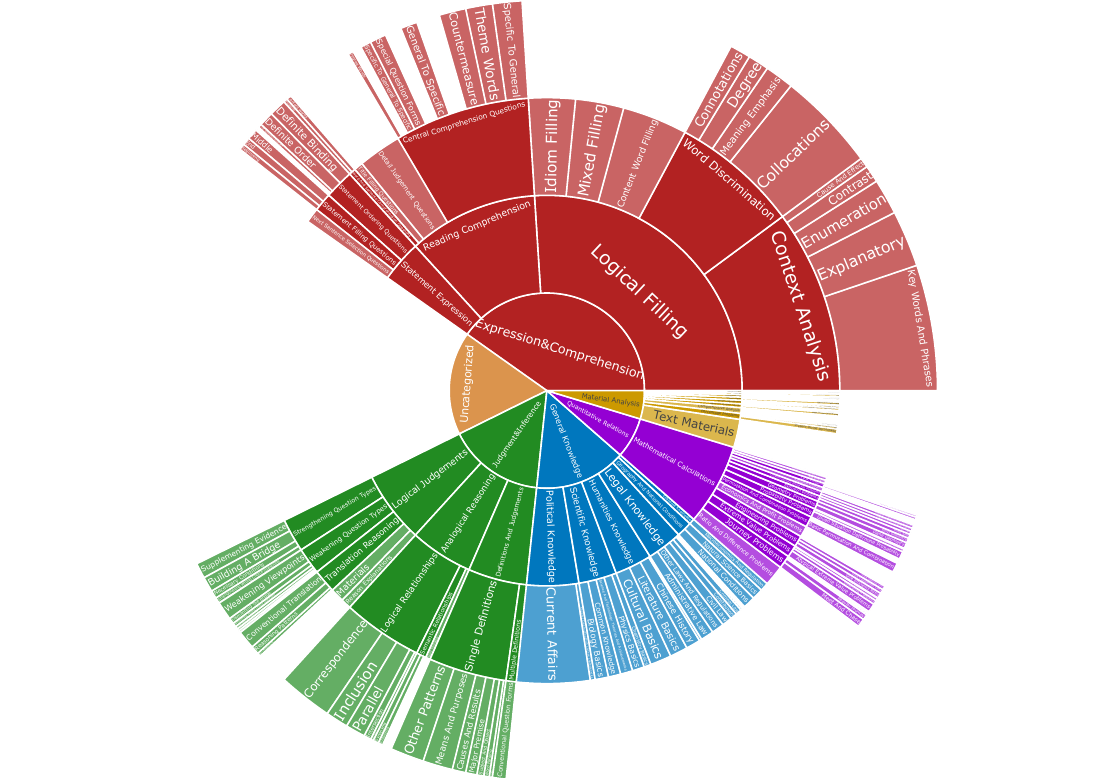}}
\caption{Keypoint Tree: The portion of the area represents the data size, including 171 keypoints and 4 levels}
\label{Keypoint Tree}
\end{figure}

The keypoint categorization can comprehensively represent the distribution of capabilities required to complete a question. These 5 major categories are identified in ANGO as the first-level keypoints.

\begin{itemize}
    
\item \textbf{Comprehension\&Expression} is the core keypoint that permeates the entire process as a fundamental ability.  

\item \textbf{Quantitative Relations} showcases computational abilities, thus reflecting the model's mathematical proficiency. 

\item \textbf{Material Analysis} represents the model's capacity to scrutinize materials, assessing the aptitude for context analysis. 

\item \textbf{Judgement\&Inference} is designed for logical structure or discernible patterns, demanding inferential ability. 

\item \textbf{General Knowledge} serves as a direct manifestation of the model's knowledge breadth and guarantees reliability.

\end{itemize}
Considering the inclusion and similarity relations among keypoints, we also constructed a Keypoint Tree consisting of 4 levels and a total of 171 nodes as shown in Figure \ref{Keypoint Tree}.

\subsubsection{Difficulty Measurement}
\begin{figure}[ht]
  \centering
  \includegraphics[width=\linewidth]{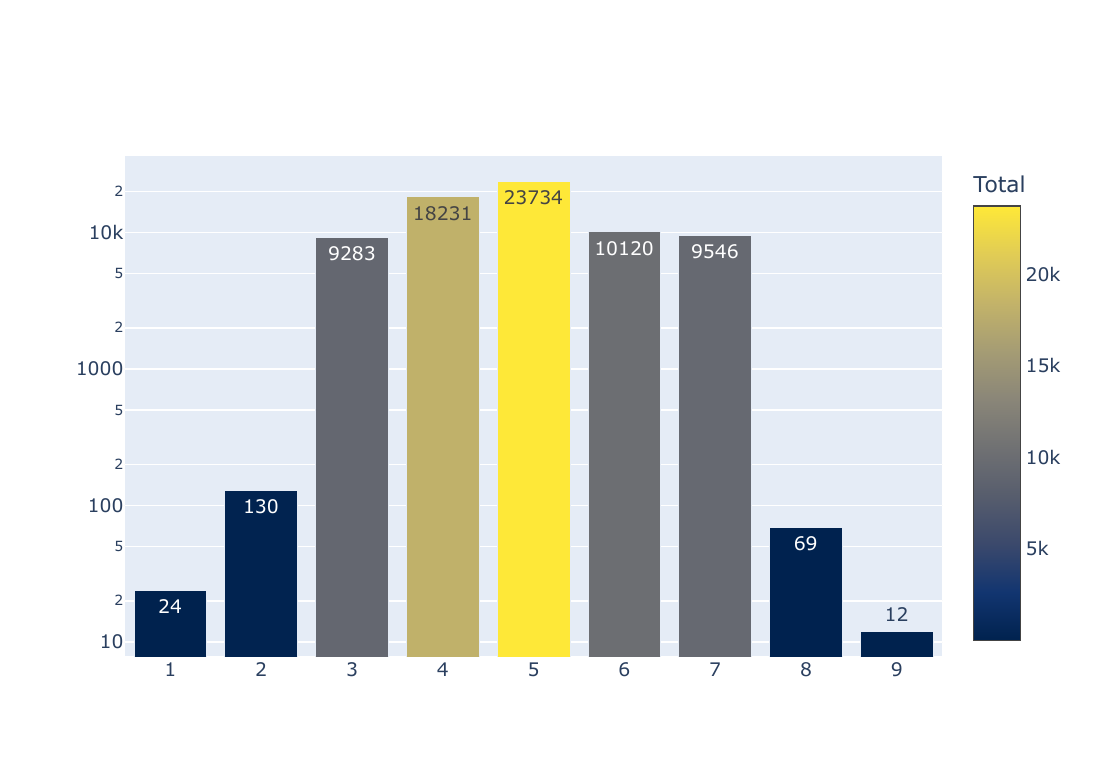}
  \caption{Difficulty Distribution: x-axis stands for difficulty level and y-axis means the question count}
  \label{Difficulty Distribution}
\end{figure}

The utilization of standardized and objective methods is crucial in measuring the difficulty of questions. Drawing upon the actual scores derived from the AAT and statistical indicators of human performance, ANGO proposes a difficulty score standard:
\begin{equation}
S(q) = \alpha S_q - \beta Acc_q - \gamma \sum_K^{K_s} Acc_K
\end{equation}

Where $S_q$ represents the score assigned to question $q$ in the actual exam, $Acc_q$ denotes the human accuracy of question $q$, $K_s$ refers to the keypoints combination of question $q$, $K$ represents the keypoint included in $K_s$, and $Acc_K$ represents the average accuracy of all questions related to $K$. Besides, the $\alpha$, $\beta$ and $\gamma$ are penalty terms.

With this standard, we divided questions in ANGO into 9 distinct levels, as shown in Figure \ref{Difficulty Distribution}, ranging from Level 1 (easiest) to Level 9 (most challenging).

\section{Sample Strategy}
\label{Sample Strategy}
Prior to commencing, it is essential to establish certain definitions to circumvent any potential misunderstandings:
\begin{itemize}
\item \textbf{$K_s$}: Refers to the combination of keypoints present in each record.
\item \textbf{$K$}: Each $K_s$ encompasses one or more keypoints, denoted as $K$.
\item \textbf{$P(K)$}: Node path from root node to target node $K$, contains a series of parent nodes $K^p$.
\item \textbf{$G(K)$}: Denotes the group of the sub nodes $K^g$ which have same parent node with $K$ in the keypoint tree.
\end{itemize}

\subsection{Few-Shot History Sample}\label{Few-Shot History Sample}
Given the unique category hierarchy in our keypoint tree architecture, coupled with a limited number of records for certain questions' $K_s$ , we designed a dynamic length example sampling strategy. This strategy ensures minimal information loss and is split into four steps. Proceeding to the next step only occurs if no available records exist in the current step.
\begin{itemize}
\item \textbf{Step 1-$K_s$:} Assign records as examples with identical $K_s$ as examples directly.

\item \textbf{Step 2-$K$:} Iterate through the $K$ within $K_s$ as a loop and sample one record associated with each $K$.

\item \textbf{Step 3-$K^p$:} Iterate through the $K$ within $K_s$ as a loop and sample one record associated with each $K^p$ in $P(K)$.

\item \textbf{Step 4-$K^g$:} Iterate through the $K$ within $K_s$ as a loop and sample one record associated with each $K^g$ in $G(K)$.
\end{itemize}

Benefits from the multi-step sampling approach, it ensures that the examples in the beginning are more closely aligned with the target question compared to those at the end.

\subsection{Testset Sample}
Since $K_s$ may contain more than one $K$ and and there’s significant overlap among different $K_s$, sampling records for each $K$ or $K_s$ will definitely lead to a imbalance result. To obtain a more even distribution, we suggest the following strategic sampling method:

\paragraph{Build $K_s$ Vectors:}Represent each $K_s$ as a vector with 171-dimensions, denoted as $V_{K}$. Each dimension in this vector corresponds to a specific $K$, $1$ denotes the presence of $K$ in $K_s$, while $0$ means absence.

\paragraph{Initialize Testset:} \label{Initialize testset step}After constructing the vectors, randomly select one record for each $K_s$ to populate the testset. Subsequently, sum the 988 $V_{K}$ to form a single vector, denoted as $V_{C}$, where each dimension represents the count of the $K$ in the current testset.

\paragraph{Insert and Update:} Firstly, define the number of records that need to be sampled for each $K$ as $N$. Then, measure the distances between $V_{C}$ and all $V_{K}$ with following distance function:

\begin{equation}
D(V_C, V_K) = \sum log_{N}V_C * \frac{V_K}{\| V_K\|_1}
\end{equation}
Here, ${\| V_K\|_1}$ is employed to normalize the weight of $K_s$, and the logarithmic transformation $log_{N}$ is applied to prevent the inhibition of some $K$ that may only be associated with $K_s$ encompassing other high-frequency $K$.

After calculation, select the $K_s$ with the minimum distance, choose a record with the difficulty level that appears the least frequently in the current testset, and insert it into the test. Concurrently, update $V_{C}$ by adding the selected $V_{K}$. 

Just repeat \textbf{Insert and Update} until each $K$ has $N$ records or there are no more records available for $K$.

\section{Evaluation}
\subsection{Metric}
We choose accuracy to be the metric for assessing the performance of our models.

For the base model(Pre-Trained stage), as there are questions with multiple correct options in our dataset, it's not feasible for us to compute the logits of varying options to derive the answer, we opted to divide the model's generated output by newline delimiters, subsequently extracting the first element as the model's answer.

For chat models(Supervised Fine Tune stage), models often failed to provide response in our expected format. To rectify this problem, we split the generated output based on various regular expressions that include Chinese and English tags, then manually vetted and captured the options as the model's answer.

\paragraph{Valuable Failure:}
\label{Valuable Failure}
Evaluating models solely based on their ability to provide correct answers does not reflect the full spectrum of human-like behavior and thinking patterns. While accuracy is important, overlooking the value in identifying common human errors may hinder progress toward building truly human-centric AI. This motivated us to propose a new evaluation metric to address this limitation. 

\textit{Human Hit} means the option predicted by the model is closest to the majority of human results. When the model identifies an option as "Most Wrong", which is the option that humans are most likely to choose incorrectly, and the human accuracy is below 50\%, it's a "Wrong Hit". Conversely, if the model gets the correct answer and the human accuracy is above 50\%, referred to as a "Correct Hit". \textit{Human Hit} is produced by the sum of these two types of Hit and serves as an approximate method to assess whether the model exhibits a tendency similar to the widespread human misconceptions found in the data.

\textit{Human Value} is obtained by iterating through all the questions in \textit{Human Hit} using the following formula.
\begin{equation}
V(q) = \sum_{i}^{n}\frac{1}{N}\left\{
\begin{array}{ll}
    Acc_q & Acc_q \geq 0.5, \\
    1 - Acc_q & Acc_q < 0.5.
\end{array}
\right.
\label{Human Value Formula}
\end{equation}
where $Acc_q$ represents Human Accuracy of the question, and $n$ denotes the count of \textit{Human Hit} and $N$ stands for total count.

While accuracy remains paramount, recognizing value in failure opens new possibilities for the model\cite{an2023learning}. \textit{Human Hit} and \textit{Human Value} aim to foster models that learn and think in a manner increasingly aligned with humanity's shared imperfections. This anthropic evaluation approach holds promise for developing AI with new features, like Generative Agents\cite{park2023generative}, better attuned to human needs, values, and experiences.

\subsection{Prompt}
\begin{figure}[ht]
\begin{center}
\centerline{\includegraphics[width=\columnwidth]{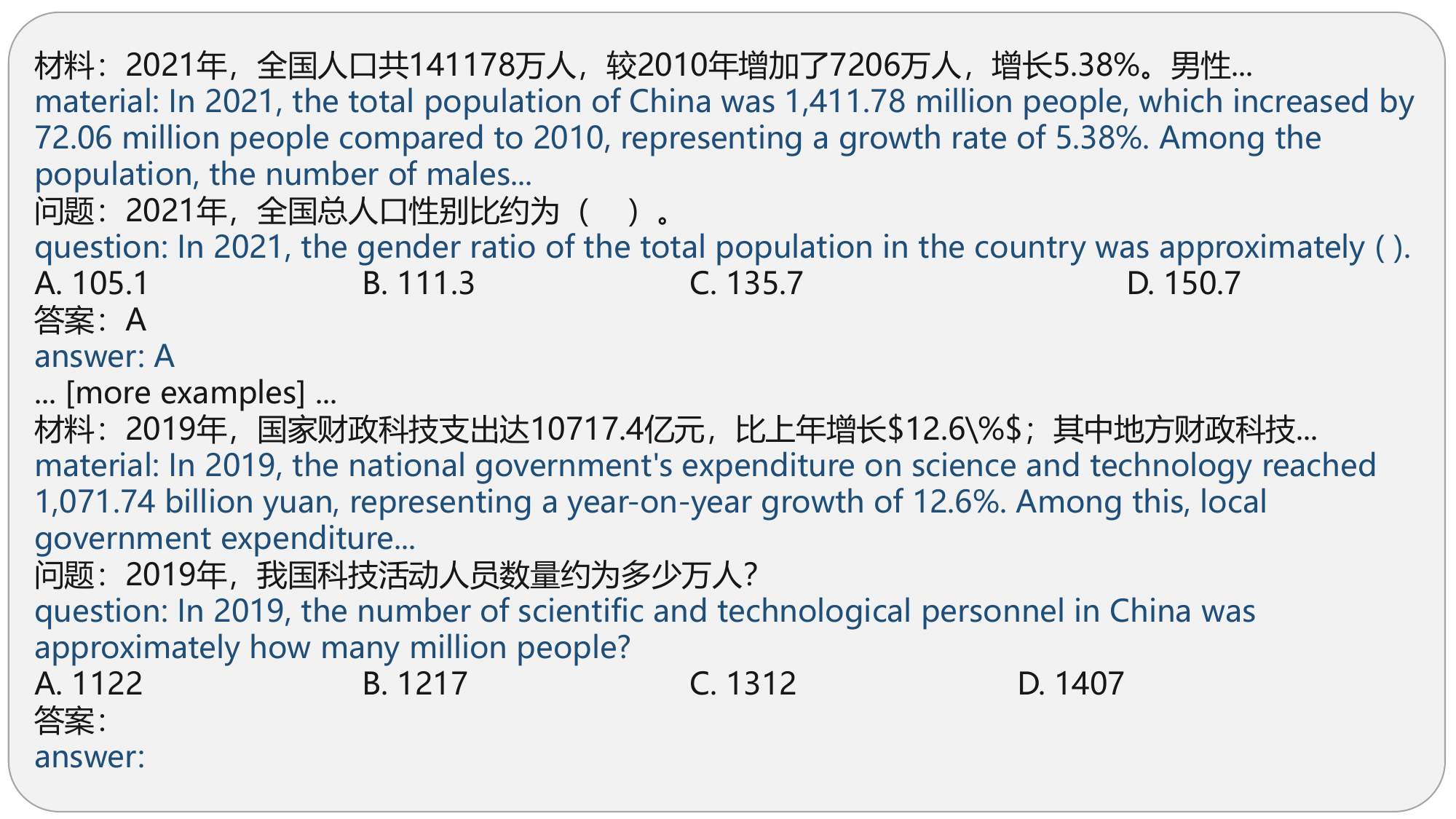}}
\caption{Prompt Example: English translations are shown below the corresponding Chinese text for better readability.}
\label{Prompt}
\end{center}
\end{figure}

Similar to the settings of other benchmarks, ANGO adopts the few-shot as an evaluation strategy with history examples sampled in Section \ref{Few-Shot History Sample}. By exposing the model to a limited number of examples during the evaluation process, we can effectively gauge its generalization and adaptability to novel scenarios. Furthermore, the few-shot evaluation strategy offers practical advantages when it comes to extracting the model's output options. Prompt example is shown in Figure \ref{Prompt}.

We have consciously chosen to minimize the incorporation of the Chain Of Thought (CoT)\cite{wei2023chainofthought} in our evaluation processes, excluding the API which remains beyond our control. This decision mainly stems from previous research\cite{huang2023ceval} which conclusively evidenced the negative impact of CoT on model performance.

When the length surpasses the limit of model, we methodically reduce the quantity of examples from the end towards the beginning. Thanks to our sample strategy (outlined in Section \ref{Sample Strategy}), the examples at the begin offer more informative content when compared to those placed at the end.

\subsection{Framework}
We devised a range of solutions to optimize the evaluation framework to address the distorting impact of data leaks and malicious intrusions on the testset, which aim to provide more authoritative and trustworthy guidance to the participating models.
\begin{itemize}
\item \textbf{Seasonal Dynamic Evaluation:} We will periodically refresh a portion of the testset and update the leaderboard so that each season has a distinct testset to prevent data leaks.

\item \textbf{Confusion of Options Order:} To prevent models from relying on spurious patterns in the order of multiple-choice options, we shuffle the order of options for each question before making predictions.

\item \textbf{Question Elimination:} By calculating questions' accuracy across models and seasons. Questions exceeding 100\% accuracy will temporarily be removed to ensure sufficient discriminability.

\end{itemize}

Together, these mechanisms enhance fairness and effectiveness compared to current leaderboards. The seasonal performance history also provides insights into model optimization trends, serving as a guide for further training.

\section{Experiment}
\subsection{Objectives}
\begin{itemize}
    \item Apply the keypoint categories of ANGO in practice, provide detailed evaluation result, and analyze the model's performance under different keypoints.
    \item Verify the difficulty grading of ANGO, provide a comparison of different grades, and demonstrate the effectiveness of difficulty classification.
    \item Test the evaluation framework of ANGO, provide an overall performance of current models, and validate the rationality of the Human Value metric.
\end{itemize}

\subsection{Models}
In light of limited computational resources, we were compelled to utilize a single A100 unit for conducting our experiments. As a result, we were obliged to select the following models as our candidates.

\textbf{GPT:} OpenAI's GPT has long been regarded as a benchmark, which has achieved SOTA in multiple domains and provides services through API. We select GPT-4\cite{openai2023gpt4} and ChatGPT\cite{openai2022chatgpt} as candidates.

\textbf{LLaMA \& LLaMA Variant:} LLaMA, which is released by Meta as an open-source LLM, allows researchers to fine-tune models for specific domains. It is the most widely adopted open-source model now. For our evaluation, we choose the LLaMA\cite{touvron2023LLaMA}, LLaMA2\cite{touvron2023LLaMA2} and some variant models that re-train on the LLaMA base model, such as ChineseLLaMA/ChineseLLaMA2\cite{cui2023efficient}, Atom\cite{flag2023atom}. 

\textbf{Chinese-Oriented:} In addition to the English-oriented LLMs mentioned above, we have also selected several LLMs that specialize in the Chinese domain. Including Baichuan/Baichuan2\cite{yang2023baichuan}, ChatGLM/ChatGLM2\cite{zeng2022glm130b}, Qwen\cite{bai2023qwen}, InternLM\cite{2023internlm}, and MOSS\cite{sun2023moss}.

\subsection{Results}
The evaluation results of ANGO are actually a multidimensional representation, and we have selected three coarse-grained distributions to showcase from it. 
\subsubsection{Keypoint Level}
\begin{figure}[htb]
  \centering
  \begin{minipage}[s]{0.45\linewidth}
      \centering
      \includegraphics[width=\linewidth]{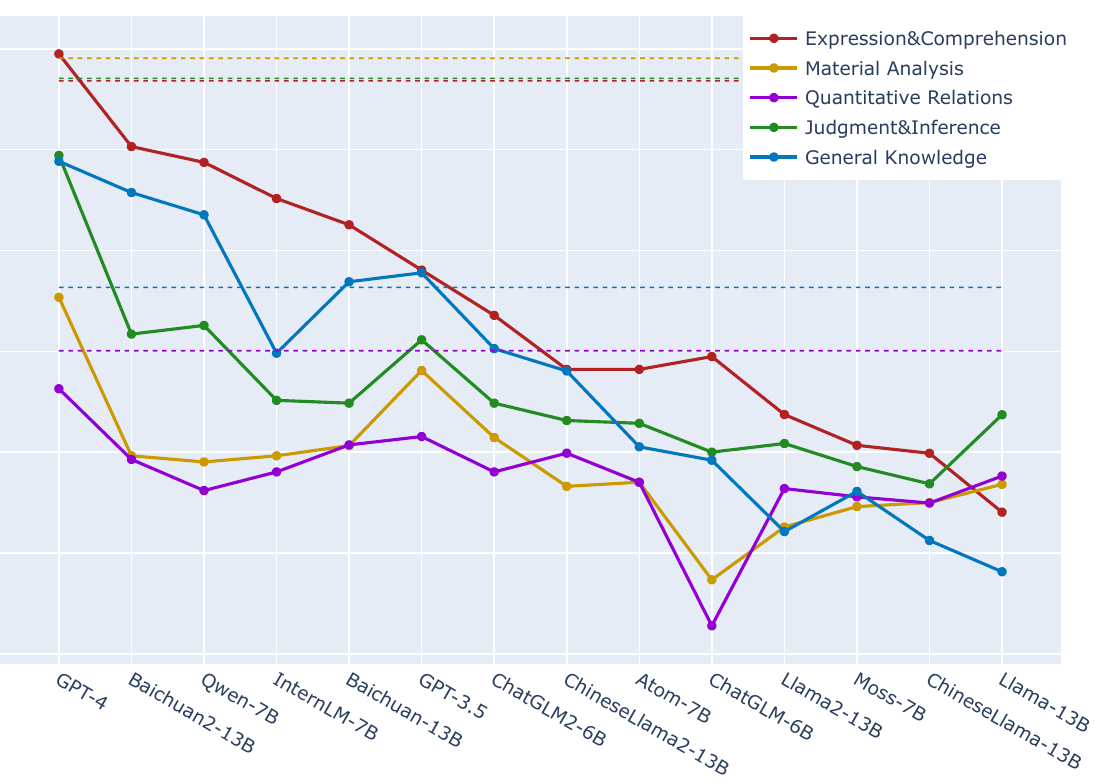}
      \caption{Performance on level 1 Keypoints in ANGO(dot line is the human performance))}
      \label{Keypoint Line}
  \end{minipage}
  \hfill
  \begin{minipage}[s]{0.45\linewidth}
    \centering
    \resizebox{\linewidth}{!}{
    \begin{tabular}{lcccccc}
    \toprule
    \multirow{2}{*}{Model}&\multirow{2}{*}{All}&Expression\&&Material&Quantitive&Judgement\%&General \\
    &&Comprehension&Analysis&Relations&Inference&Knowledge \\
    \midrule
    Total&{3863}&{2296}&{496}&{485}&{350}&{226} \\
    \midrule   Human&\textbf{62.54}&66.82&\textbf{69.07}&\textbf{40.06}&\textbf{67.08}&46.34\\
    \midrule
    GPT-4&60.67&\textbf{69.5}&45.36&36.29&59.43&\textbf{58.85}\\
    Baichuan2-13B&50.52&60.31&29.64&29.28&41.71&55.75\\
    Qwen-7B&49.07&58.74&29.03&26.19&42.57&53.54\\
    Baichuan2-7B&46.22&54.38&30.24&28.25&38.29&50.0\\
    InternLM-7B&45.7&55.16&29.64&28.04&35.14&39.82\\
    Baichuan-13B&44.98&52.55&30.65&30.72&34.86&46.9\\
    ChatGPT&44.02&48.06&38.1&31.55&41.14&47.79\\
    ChatGLM2-6B&39.0&43.57&31.45&28.04&34.86&40.27\\
    Baichuan-7B&38.01&43.92&30.85&26.6&30.29&30.09\\
    ChineseLLaMA2-13B&35.19&38.21&26.61&29.9&33.14&38.05\\
    Atom-7B&34.41&38.21&27.02&27.01&32.86&30.53\\
    ChatGLM-6B&31.75&39.48&17.34&12.78&30.0&29.2\\
    LLaMA2-13B&30.42&33.73&22.58&26.39&30.86&22.12\\
    ChineseLLaMA2-7B&29.98&30.02&31.45&29.69&30.86&26.11\\
    Moss-7B&28.85&30.68&24.6&25.57&28.57&26.11\\
    ChineseLLaMA-13B&27.86&29.89&25.0&24.95&26.86&21.24\\
    LLaMA2-7B&27.55&28.28&27.22&27.01&27.43&22.12\\
    LLaMA-13B&25.43&24.05&26.81&27.63&33.71&18.14\\
    LLaMA-7B&25.01&26.8&23.59&21.24&24.0&19.91\\
    \bottomrule
    \end{tabular}
    }
    \captionof{table}{Result On Keypoint Tree Level-1: Because a single question may contain more than one keypoint, the total number of keypoint count is higher than the question count}
    \label{Keypoint Level}
  \end{minipage}
\end{figure}
Figure \ref{Keypoint Line} clearly reveals that models are able to achieve higher accuracy levels in the \textit{Expression\&Comprehension} and \textit{General Knowledge}. Because the question formats and content in these two categories tend to be more inclined towards simple content completion, which is relatively well-suited to the pretraining process of language models\cite{brown2020language}.

In the \textit{Judgement\&Inference}, model performances closely align with the average accuracy. Dive deeper, we can see some conventional logical problems are well-handled by the models\cite{chang2023survey}. However, there is still a subset of questions in this part that involve pattern-based problems, where models often struggle due to their inability to comprehend the problem's context.

Mathematics is widely recognized as a challenging task for LLM\cite{kaddour2023challenges}. This challenge persists as all models demonstrate significantly lower accuracy levels in the domain of \textit{Quantitative Relations}. In contrast to existing benchmarks, the format of mathematical questions in ANGO tends to rely more heavily on extensive linguistic content, placing a greater emphasis on the combined language and computational capabilities of the models.

\textit{Material Analysis} is a crucial keypoint of ANGO, characterized by the longer context resulting from the combination of material and question. Due to the extensive context, models often face challenges of attention dispersion, leading to hallucination\cite{Ji_2023}. Furthermore, this category always involves different abilities, such as analysis, reasoning, and computation. Therefore, the accuracy in this section is significantly much lower than in humans.

\paragraph{Why ChatGPT fail?}
Regarding the previously mentioned underperformance of ChatGPT, Table \ref{Keypoint Level} and Figure \ref{Keypoint Line} demonstrates that ChatGPT can achieve better performance in challenging categories such as \textit{Material Analysis} and \textit{Quantitative Relations} by leveraging specialized tools and enhanced capabilities.

In terms of \textit{Expression\&Comprehension} and \textit{General Knowledge}, which represent a significant proportion. Models specifically trained for Chinese achieve better performance level than others. This status can be attributed to differences in the distribution of language features and locally specific knowledge, causing ChatGPT to struggle.

Furthermore, we observed that when faced with challenging questions containing additional information embedded within specific structures and sentence patterns, ChatGPT tends to generate insufficient conditions or unable-to-answer responses rather than accurately selecting the most relevant answer option, which is primarily due to the overreliance on RLHF,  \cite{chen2023chatgpts}.

\subsubsection{Difficulty Level}
\begin{figure}[htb]
  \centering
  \begin{minipage}[s]{0.45\linewidth}
      \centering
      \includegraphics[width=\linewidth]{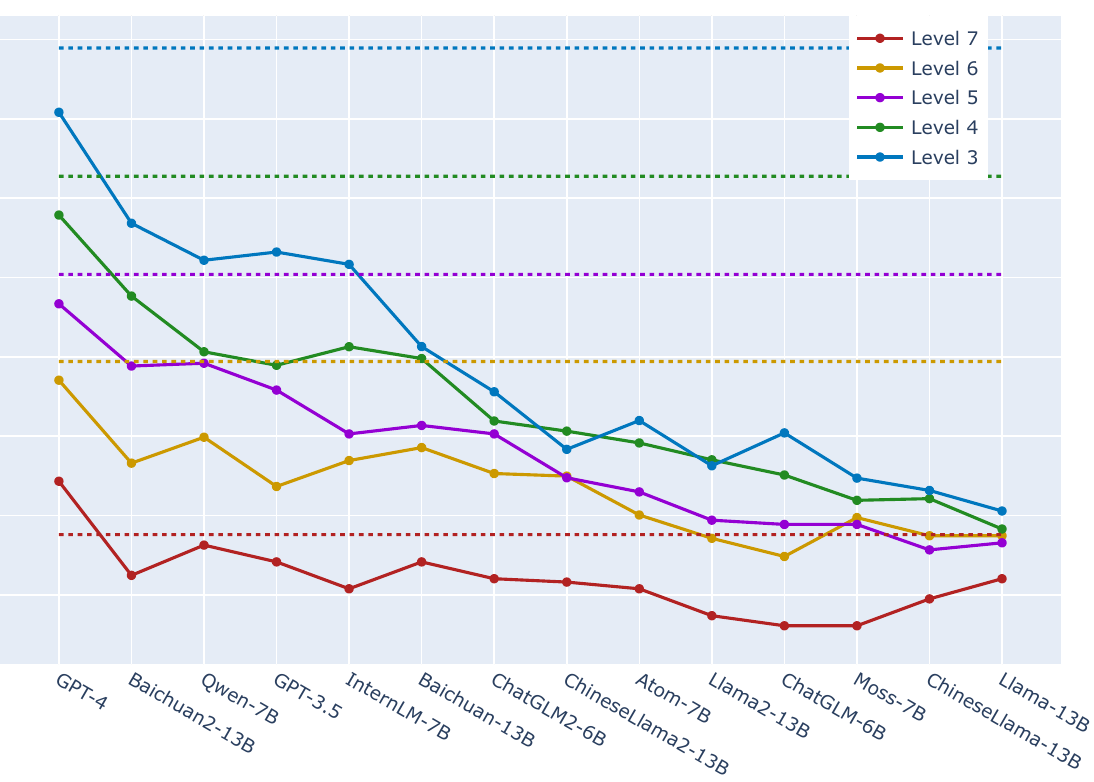}
      \caption{Accuracy on different different difficulty levels(dot line is the human performance)}
      \label{Difficulty Line}
  \end{minipage}
  \hfill
  \begin{minipage}[s]{0.45\linewidth}
    \centering
    \resizebox{\columnwidth}{!}{
    \begin{tabular}{lccccccc}
    \toprule
    Difficult&{All}&{Level 7}&{Level 6}&{Level 5}&{Level 4}&{Level 3}&{Level 2} \\
    \midrule
    Total&{1768}&{236}&{306}&{561}&{470}&{193}&{2} \\
    \midrule
    Human&\textbf{62.54}&27.59&\textbf{49.40}&\textbf{60.4}&\textbf{72.76}&\textbf{88.92}&96.78\\
    \midrule
    GPT-4&57.69&\textbf{34.32}&47.06&56.68&67.87&80.83&\textbf{100.0}\\
    Baichuan2-13B&47.51&22.46&36.6&48.84&57.66&66.84&50.0\\
    Qwen-7B&46.38&26.27&39.87&49.2&50.64&62.18&100.0\\
    Baichuan2-7B&43.27&24.15&35.62&43.32&51.7&57.51&100.0\\
    ChatGPT&43.21&23.73&33.66&44.92&49.36&61.66&100.0\\
    InternLM-7B&42.36&20.76&36.93&40.29&51.28&61.66&50.0\\
    Baichuan-13B&41.97&24.15&38.56&41.35&49.79&51.3&100.0\\
    ChatGLM2-6B&38.01&22.03&35.29&40.29&41.91&45.6&50.0\\
    Baichuan-7B&35.35&17.8&33.99&34.58&40.43&48.19&100.0\\
    ChineseLLaMA2-13B&35.07&21.61&34.97&34.76&40.64&38.34&100.0\\
    Atom-7B&33.48&20.76&30.07&32.98&39.15&41.97&50.0\\
    ChatGLM-6B&30.32&16.53&24.84&30.66&35.96&40.93&50.0\\
    ChineseLLaMA2-7B&30.2&18.22&30.72&29.95&34.68&33.68&50.0\\
    LLaMA2-13B&30.2&17.37&27.12&29.41&37.02&36.27&50.0\\
    Moss-7B&28.73&16.1&29.74&28.88&31.91&34.72&0.0\\
    ChineseLLaMA-13B&27.71&19.49&27.45&25.67&32.13&33.16&50.0\\
    LLaMA2-7B&27.43&18.64&28.1&27.09&30.64&30.05&50.0\\
    LLaMA-13B&27.04&22.03&27.45&26.56&28.3&30.57&50.0\\
    LLaMA-7B&24.66&13.98&24.84&24.24&28.94&28.5&0.0\\
    \bottomrule
    \end{tabular}
    }
    \captionof{table}{Result on Different Difficulty}
    \label{Difficulty Level}
  \end{minipage}
\end{figure}

As evident from the Figure \ref{Difficulty Line}, under the difficulty standards set by ANGO, the performance of the models indeed exhibits a strong negative correlation, indicating that higher difficulty levels are associated with lower accuracy. Furthermore, the model that performs well on average does not necessarily perform well across all difficulty levels, but the overall ranking distribution still aligns with the relationship to the average performance, demonstrating the differentiation across difficulty levels in ANGO. All of these findings confirm that the difficulty standard in ANGO is a reasonable and effective quantitative measure and provides a systematic and objective approach to assessing question difficulty in evaluation. 

Besides, we also found that models usually work well on high difficulty level questions compared to real human. 

Based on the analysis of keypoint performance, it is clear that the LLM's robust "memory" capabilities equip it to effectively address challenging questions within specific domain of knowledge. In contrast, real human generally possess a restricted comprehension across diverse subject areas, yet they excel in solution with limited information. As a result, the model's superior performance in high difficulty level questions, which belong to \textit{Expression\&Comprehension} and \textit{General Knowledge}, and this empowers the model to outperform humans in tackling intricate problems.

\subsubsection{Question Level}
\begin{figure}[htb]
  \centering
  \begin{minipage}[s]{0.45\linewidth}
      \centering
      \includegraphics[width=\linewidth]{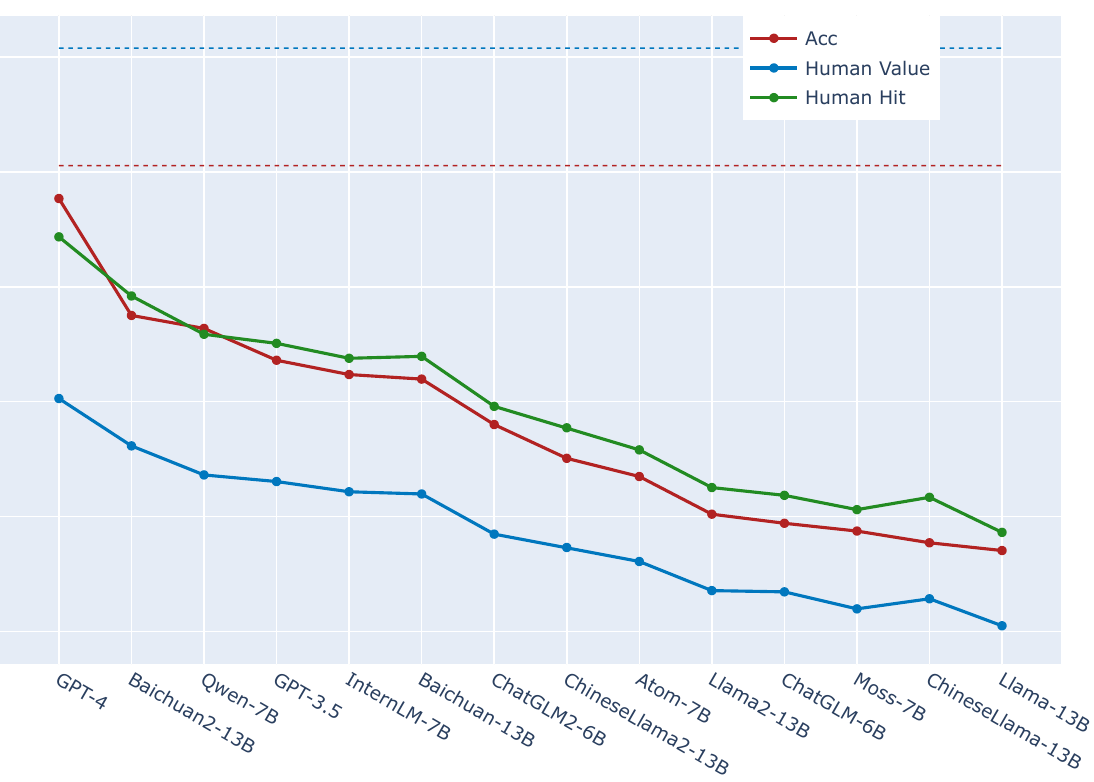}
      \caption{Acc\&Human Value: Dot line is the human performance, ignore the Human Hit of Human(because it's 100\%)}
      \label{Human Value}
  \end{minipage}
  \hfill
  \begin{minipage}[s]{0.45\linewidth}
    \centering
    \resizebox{\columnwidth}{!}{
    \begin{tabular}{lcccccc}
    \toprule
    \multirow{2}{*}{Model}&\multirow{2}{*}{Acc \%}&Human&Human&Human&\multirow{2}{*}{Correct}&\multirow{2}{*}{Total} \\
    &&Value \%&Hit \%&Hit&& \\
    \midrule
    Human&60.5594&70.797&100&1768&1208&1768\\
    \midrule
    GPT-4&57.6923&40.2748&54.3552&961&1020&1768\\
    Baichuan2-13B&47.5113&36.1515&49.2081&870&840&1768\\
    Qwen-7B&46.3801&33.6215&45.871&811&820&1768\\
    ChatGPT&43.6086&33.0455&45.0792&797&771&1768\\
    Baichuan2-7B&43.2692&32.6535&44.5701&788&765&1768\\
    InternLM-7B&42.3643&32.1553&43.7783&774&749&1768\\
    Baichuan-13B&41.9683&31.9635&43.948&777&742&1768\\
    ChatGLM2-6B&38.009&28.4656&39.5928&700&672&1768\\
    Baichuan-7B&35.3507&28.2665&38.9706&689&625&1768\\
    ChineseLLaMA2-13B&35.0679&27.3031&37.7262&667&620&1768\\
    Atom-7B&33.4842&26.0837&35.8032&633&592&1768\\
    ChineseLLaMA2-7B&30.2036&23.3305&32.2964&571&534&1768\\
    LLaMA2-13B&30.2036&23.5473&32.5226&575&534&1768\\
    ChatGLM-6B&29.4118&23.4395&31.8439&563&520&1768\\
    Moss-7B&28.733&21.9575&30.5995&541&508&1768\\
    ChineseLLaMA-13B&27.7149&22.8383&31.6742&560&490&1768\\
    LLaMA2-7B&27.4321&21.6132&29.8643&528&485&1768\\
    LLaMA-13B&27.0362&20.485&28.6199&506&478&1768\\
    LLaMA-7B&24.6606&20.6606&29.0724&514&436&1768\\
    \bottomrule
    \end{tabular}
    }
    \captionof{table}{Question Level Result: ”Acc" of human is computed by averaging the human accuracy, and the 'Hit' of human is determined by the number of questions where the "Acc" of human is above 50\%.}
    \label{Question Level}
  \end{minipage}
\end{figure}

Based on the evaluation results from ANGO, as shown in Table \ref{Question Level}, the majority of participating models performed lower in terms of accuracy compared to existing Chinese benchmarks. This finding indicates that ANGO presents greater challenges for models, thereby enabling more effective identification and performance evaluation.

It is noteworthy that, as shown the ranking details, all English-oriented models performed poorly, and even ChatGPT being outperformed by models with significantly fewer parameters. This outcome can be attributed to ANGO's emphasis on specific Chinese language features.

\paragraph{Human Value}
From Figure \ref{Human Value}, there is a positive correlation between Acc and Human Value, indicating that Human Value can indeed reflect the model's capabilities rather than being arbitrarily fabricated. However, fluctuations are still observed between them within certain ranges. This suggests that Human Value is not a subset or a new expression form of Acc, but rather a distinct indicator. Overall, these quantitative results preliminarily validate that the concept of Human Value captures meaningful signals from a specific evaluative standpoint, affirming that assigning a certain value to model errors can indeed serve as a novel perspective for assessing models.

Based on the results, it can be concluded that models with higher Human Value tend to generate responses that resemble human-like outputs rather than strictly correct answers during conversations. This is because most real humans have limited knowledge in specific domains, and it is precisely this limitation that nurtures specific values and ways of thinking, ultimately shaping different human personalities and traits, which we refer to as "human nature". When the goal is to train AI systems that exhibit "human-like" rather than being "omniscient", incorporating Human Value as a guiding principle may be beneficial.

\section{Related Work}
The field of English language assessment began its exploration of multiple-choice question benchmarks with the pioneering work of MMLU\cite{hendrycks2021measuring}. MMLU constructed a comprehensive dataset comprising 57 subjects with varying levels of difficulty. This dataset was created by collecting multiple-choice questions in the form of question-answer pairs. Another significant contribution to the evaluation of model truthfulness in question-answering tasks was introduced by TruthfulQA\cite{lin2022truthfulqa}. TruthfulQA transformed traditional question-answering tasks into a multiple-choice format, enabling the assessment of model truthfulness. BIG-Bench\cite{srivastava2023beyond} took the benchmarking of models to a new level by expanding the classification dimension from two-digit to three-digit levels. It also maintained the dataset dynamically, ensuring its relevance and adaptability.

In the Chinese language domain, CMMLU\cite{li2023cmmlu} and MMCU\cite{zeng2023measuring} constructed benchmarks inspired by MMLU. These benchmarks were divided based on examination subjects and sources, providing valuable resources for evaluating Chinese language models. AGI-Eval\cite{zhong2023agieval} approached benchmarking from a different perspective and optimized it for Chinese characteristics. It created a human-centric benchmark by evaluating model performance based on real human performance, aiming to capture the nuances of Chinese language understanding. Following closely, the release of C-Eval\cite{huang2023ceval} introduced new possibilities for Chinese benchmarks. C-Eval incorporated sophisticated reasoning problems and made efforts to address the issue of data leaks, providing a more comprehensive evaluation platform.

In comparison to existing Chinese domain datasets, ANGO stands out for several distinct features: (1) ANGO includes more complex problems that require models to combine different abilities (2) ANGO incorporates measurements of problem difficulty based on real human performance, using a continuous quantitative metric instead of discrete judgments. (3) Benefited by carefully designed and optimized sampling strategies, the ANGO benchmark enables rapid iteration and minimizes the potential for malicious hacks and data leaks once the leaderboard is launched.

\section{Conclusion}
The wide coverage and reliability of the ANGO make it a valuable resource that fosters innovation and progress in NLP. By conducting systematic analysis and application of ANGO, researchers can gain a profound understanding of the performance and capabilities of models, thereby providing guidance for model design and improvement. And the unique feature of the ANGO enable developers to better evaluate and enhance their own models. We encourage researchers and developers to actively explore and utilize the ANGO, thereby driving further development and innovation in the field of LLM.

\bibliographystyle{unsrt}  
\bibliography{ango}  

\newpage
\appendix
\section{Keypoint Chinese-English Mapping}
\subsection{Expression\&Comprehension}
\resizebox{\columnwidth}{!}{%
\begin{CJK*}{UTF8}{gbsn}
\begin{tabular}{|c|c|c|c|c|c|c|c|}
    \toprule
    Level-1 Chinese&Level-1 English&Level-2 Chinese&Level-2 English&Level-3 Chinese&Level-3 English&Level-4 Chinese&Level-4 English \\
    \midrule
    言语理解与表达&Expression\&Comprehension&&&&&&\\
    \midrule
    言语理解与表达&Expression\&Comprehension&阅读理解&Reading Comprehension&&&&\\
    \midrule
    言语理解与表达&Expression\&Comprehension&阅读理解&Reading Comprehension&中心理解题&Central Comprehension Questions&&\\
    \midrule
    言语理解与表达&Expression\&Comprehension&阅读理解&Reading Comprehension&中心理解题&Central Comprehension Questions&关联词-转折&Contrast\\
    \midrule
    言语理解与表达&Expression\&Comprehension&阅读理解&Reading Comprehension&中心理解题&Central Comprehension Questions&关联词-因果&Cause And Effect\\
    \midrule
    言语理解与表达&Expression\&Comprehension&阅读理解&Reading Comprehension&中心理解题&Central Comprehension Questions&关联词-对策&Countermeasure\\
    \midrule
    言语理解与表达&Expression\&Comprehension&阅读理解&Reading Comprehension&中心理解题&Central Comprehension Questions&关联词-并列&Enumeration\\
    \midrule
    言语理解与表达&Expression\&Comprehension&阅读理解&Reading Comprehension&中心理解题&Central Comprehension Questions&主题词&Theme Words\\
    \midrule
    言语理解与表达&Expression\&Comprehension&阅读理解&Reading Comprehension&中心理解题&Central Comprehension Questions&程度词&Degree Words\\
    \midrule
    言语理解与表达&Expression\&Comprehension&阅读理解&Reading Comprehension&中心理解题&Central Comprehension Questions&行文脉络-总分&General To Specific\\
    \midrule
    言语理解与表达&Expression\&Comprehension&阅读理解&Reading Comprehension&中心理解题&Central Comprehension Questions&行文脉络-分总&Specific To General\\
    \midrule
    言语理解与表达&Expression\&Comprehension&阅读理解&Reading Comprehension&中心理解题&Central Comprehension Questions&行文脉络-分总分&Specific To General To Specific\\
    \midrule
    言语理解与表达&Expression\&Comprehension&阅读理解&Reading Comprehension&中心理解题&Central Comprehension Questions&特殊问法&Special Question Forms\\
    \midrule
    言语理解与表达&Expression\&Comprehension&阅读理解&Reading Comprehension&细节判断题&Detail Judgement Questions&&\\
    \midrule
    言语理解与表达&Expression\&Comprehension&阅读理解&Reading Comprehension&词句理解题&Word And Sentence Comprehension Questions&&\\
    \midrule
    言语理解与表达&Expression\&Comprehension&阅读理解&Reading Comprehension&词句理解题&Word And Sentence Comprehension Questions&实词&Content Words\\
    \midrule
    言语理解与表达&Expression\&Comprehension&阅读理解&Reading Comprehension&词句理解题&Word And Sentence Comprehension Questions&代词&Pronouns\\
    \midrule
    言语理解与表达&Expression\&Comprehension&阅读理解&Reading Comprehension&标题填入题&Title Filling Questions&&\\
    \midrule
    言语理解与表达&Expression\&Comprehension&语句表达&Statement Expression&&&&\\
    \midrule
    言语理解与表达&Expression\&Comprehension&语句表达&Statement Expression&语句排序题&Statement Ordering Questions&&\\
    \midrule
    言语理解与表达&Expression\&Comprehension&语句表达&Statement Expression&语句排序题&Statement Ordering Questions&首句特征&First Sentence Features\\
    \midrule
    言语理解与表达&Expression\&Comprehension&语句表达&Statement Expression&语句排序题&Statement Ordering Questions&非首句特征&Non-First Sentence Features\\
    \midrule
    言语理解与表达&Expression\&Comprehension&语句表达&Statement Expression&语句排序题&Statement Ordering Questions&确定捆绑&Definite Binding\\
    \midrule
    言语理解与表达&Expression\&Comprehension&语句表达&Statement Expression&语句排序题&Statement Ordering Questions&确定顺序&Definite Order\\
    \midrule
    言语理解与表达&Expression\&Comprehension&语句表达&Statement Expression&语句排序题&Statement Ordering Questions&尾句特征&Ending Sentence Features\\
    \midrule
    言语理解与表达&Expression\&Comprehension&语句表达&Statement Expression&语句填空题&Statement Filling Questions&&\\
    \midrule
    言语理解与表达&Expression\&Comprehension&语句表达&Statement Expression&语句填空题&Statement Filling Questions&开头&Beginning\\
    \midrule
    言语理解与表达&Expression\&Comprehension&语句表达&Statement Expression&语句填空题&Statement Filling Questions&中间&Middle\\
    \midrule
    言语理解与表达&Expression\&Comprehension&语句表达&Statement Expression&语句填空题&Statement Filling Questions&结尾&End\\
    \midrule
    言语理解与表达&Expression\&Comprehension&语句表达&Statement Expression&接语选择题&Next Sentence Selection Questions&&\\
    \midrule
    言语理解与表达&Expression\&Comprehension&逻辑填空&Logical Filling&&&&\\
    \midrule
    言语理解与表达&Expression\&Comprehension&逻辑填空&Logical Filling&实词填空&Content Word Filling&&\\
    \midrule
    言语理解与表达&Expression\&Comprehension&逻辑填空&Logical Filling&成语填空&Idiom Filling&&\\
    \midrule
    言语理解与表达&Expression\&Comprehension&逻辑填空&Logical Filling&混搭填空&Mixed Filling&&\\
    \midrule
    言语理解与表达&Expression\&Comprehension&逻辑填空&Logical Filling&词的辨析&Word Discrimination&&\\
    \midrule
    言语理解与表达&Expression\&Comprehension&逻辑填空&Logical Filling&词的辨析&Word Discrimination&词的辨析-词义侧重&Meaning Emphasis\\
    \midrule
    言语理解与表达&Expression\&Comprehension&逻辑填空&Logical Filling&词的辨析&Word Discrimination&词的辨析-固定搭配&Collocations\\
    \midrule
    言语理解与表达&Expression\&Comprehension&逻辑填空&Logical Filling&词的辨析&Word Discrimination&词的辨析-感情色彩&Connotations\\
    \midrule
    言语理解与表达&Expression\&Comprehension&逻辑填空&Logical Filling&词的辨析&Word Discrimination&词的辨析-程度轻重&Degree\\
    \midrule
    言语理解与表达&Expression\&Comprehension&逻辑填空&Logical Filling&语境分析&Context Analysis&&\\
    \midrule
    言语理解与表达&Expression\&Comprehension&逻辑填空&Logical Filling&语境分析&Context Analysis&关联关系-转折关系&Contrast\\
    \midrule
    言语理解与表达&Expression\&Comprehension&逻辑填空&Logical Filling&语境分析&Context Analysis&关联关系-因果关系&Cause And Effect\\
    \midrule
    言语理解与表达&Expression\&Comprehension&逻辑填空&Logical Filling&语境分析&Context Analysis&关联关系-并列关系&Enumeration\\
    \midrule
    言语理解与表达&Expression\&Comprehension&逻辑填空&Logical Filling&语境分析&Context Analysis&对应关系-解释类对应&Explanatory\\
    \midrule
    言语理解与表达&Expression\&Comprehension&逻辑填空&Logical Filling&语境分析&Context Analysis&对应关系-重点词句对应&Key Words And Phrases\\
    \bottomrule
\end{tabular}
\end{CJK*}
}
\vfill\null

\subsection{Quantitative Relations}
\resizebox{\columnwidth}{!}{%
\begin{CJK*}{UTF8}{gbsn}
\begin{tabular}{|c|c|c|c|c|c|c|c|}
    \toprule
    Level-1 Chinese&Level-1 English&Level-2 Chinese&Level-2 English&Level-3 Chinese&Level-3 English&Level-4 Chinese&Level-4 English \\
    \midrule
    数量关系&Quantitative Relations&&&&&&\\
    \midrule
    数量关系&Quantitative Relations&数学运算&Mathematical Calculations&&&&\\
    \midrule
    数量关系&Quantitative Relations&数学运算&Mathematical Calculations&工程问题&Engineering Problems&&\\
    \midrule
    数量关系&Quantitative Relations&数学运算&Mathematical Calculations&工程问题&Engineering Problems&给完工时间型&Give Completion Time\\
    \midrule
    数量关系&Quantitative Relations&数学运算&Mathematical Calculations&工程问题&Engineering Problems&给效率比例型&Give Efficiency Ratio\\
    \midrule
    数量关系&Quantitative Relations&数学运算&Mathematical Calculations&工程问题&Engineering Problems&给具体单位型&Give Specific Unit\\
    \midrule
    数量关系&Quantitative Relations&数学运算&Mathematical Calculations&工程问题&Engineering Problems&工程问题-其他&Other\\
    \midrule
    数量关系&Quantitative Relations&数学运算&Mathematical Calculations&最值问题&Extreme Value Problems&&\\
    \midrule
    数量关系&Quantitative Relations&数学运算&Mathematical Calculations&最值问题&Extreme Value Problems&非典型最值问题&Atypical Extreme Value Problems\\
    \midrule
    数量关系&Quantitative Relations&数学运算&Mathematical Calculations&最值问题&Extreme Value Problems&构造数列&Constructing A Sequence\\
    \midrule
    数量关系&Quantitative Relations&数学运算&Mathematical Calculations&最值问题&Extreme Value Problems&最不利构造&Most Unfavorable Construction\\
    \midrule
    数量关系&Quantitative Relations&数学运算&Mathematical Calculations&最值问题&Extreme Value Problems&多集合反向构造&Reverse Construction Of Multiple Sets\\
    \midrule
    数量关系&Quantitative Relations&数学运算&Mathematical Calculations&年龄问题&Age Problems&&\\
    \midrule
    数量关系&Quantitative Relations&数学运算&Mathematical Calculations&和差倍比问题&Ratio And Difference Problems&&\\
    \midrule
    数量关系&Quantitative Relations&数学运算&Mathematical Calculations&周期问题&Cyclic Problems&&\\
    \midrule
    数量关系&Quantitative Relations&数学运算&Mathematical Calculations&周期问题&Cyclic Problems&周期相遇问题&Cyclic Encounter Problems\\
    \midrule
    数量关系&Quantitative Relations&数学运算&Mathematical Calculations&周期问题&Cyclic Problems&周期余数问题&Cyclic Remainder Problems\\
    \midrule
    数量关系&Quantitative Relations&数学运算&Mathematical Calculations&周期问题&Cyclic Problems&周期问题-其他&Other\\
    \midrule
    数量关系&Quantitative Relations&数学运算&Mathematical Calculations&数列问题&Sequence Problems&&\\
    \midrule
    数量关系&Quantitative Relations&数学运算&Mathematical Calculations&行程问题&Journey Problems&&\\
    \midrule
    数量关系&Quantitative Relations&数学运算&Mathematical Calculations&行程问题&Journey Problems&火车过桥&Train Crossing Bridge\\
    \midrule
    数量关系&Quantitative Relations&数学运算&Mathematical Calculations&行程问题&Journey Problems&平均速度&Average Speed\\
    \midrule
    数量关系&Quantitative Relations&数学运算&Mathematical Calculations&行程问题&Journey Problems&普通行程&Regular Journey\\
    \midrule
    数量关系&Quantitative Relations&数学运算&Mathematical Calculations&行程问题&Journey Problems&相遇追及&Meet And Chase\\
    \midrule
    数量关系&Quantitative Relations&数学运算&Mathematical Calculations&行程问题&Journey Problems&流水行船&Stream Flow Ship\\
    \midrule
    数量关系&Quantitative Relations&数学运算&Mathematical Calculations&行程问题&Journey Problems&行程问题-其他&Other\\
    \midrule
    数量关系&Quantitative Relations&数学运算&Mathematical Calculations&几何问题&Geometry Problems&&\\
    \midrule
    数量关系&Quantitative Relations&数学运算&Mathematical Calculations&几何问题&Geometry Problems&平面几何&Plane Geometry\\
    \midrule
    数量关系&Quantitative Relations&数学运算&Mathematical Calculations&几何问题&Geometry Problems&立体几何&Solid Geometry\\
    \midrule
    数量关系&Quantitative Relations&数学运算&Mathematical Calculations&容斥原理问题&Exclusion Problems&&\\
    \midrule
    数量关系&Quantitative Relations&数学运算&Mathematical Calculations&容斥原理问题&Exclusion Problems&两集合&Two Sets\\
    \midrule
    数量关系&Quantitative Relations&数学运算&Mathematical Calculations&容斥原理问题&Exclusion Problems&三集合&Three Sets\\
    \midrule
    数量关系&Quantitative Relations&数学运算&Mathematical Calculations&排列组合问题&Permutation And Combination Problems&&\\
    \midrule
    数量关系&Quantitative Relations&数学运算&Mathematical Calculations&排列组合问题&Permutation And Combination Problems&基础排列组合&Basic Permutation And Combination\\
    \midrule
    数量关系&Quantitative Relations&数学运算&Mathematical Calculations&排列组合问题&Permutation And Combination Problems&相邻问题&Adjacent Issues\\
    \midrule
    数量关系&Quantitative Relations&数学运算&Mathematical Calculations&排列组合问题&Permutation And Combination Problems&不相邻问题&Non-Adjacent Issues\\
    \midrule
    数量关系&Quantitative Relations&数学运算&Mathematical Calculations&排列组合问题&Permutation And Combination Problems&同素分堆问题&Issues With Same Factors\\
    \midrule
    数量关系&Quantitative Relations&数学运算&Mathematical Calculations&排列组合问题&Permutation And Combination Problems&环形排列问题&Circular Arrangement Issues\\
    \midrule
    数量关系&Quantitative Relations&数学运算&Mathematical Calculations&排列组合问题&Permutation And Combination Problems&错位排列&Dislocation Arrangement\\
    \midrule
    数量关系&Quantitative Relations&数学运算&Mathematical Calculations&排列组合问题&Permutation And Combination Problems&排列组合问题-其他&Other\\
    \midrule
    数量关系&Quantitative Relations&数学运算&Mathematical Calculations&概率问题&Probability Problems&&\\
    \midrule
    数量关系&Quantitative Relations&数学运算&Mathematical Calculations&概率问题&Probability Problems&给情况求概率&Given Situation Determine Probability\\
    \midrule
    数量关系&Quantitative Relations&数学运算&Mathematical Calculations&概率问题&Probability Problems&给概率求概率&Given Probability Determine Probability\\
    \midrule
    数量关系&Quantitative Relations&数学运算&Mathematical Calculations&概率问题&Probability Problems&概率问题-其他&Other\\
    \midrule
    数量关系&Quantitative Relations&数学运算&Mathematical Calculations&经济利润问题&Economics And Profit Problems&&\\
    \midrule
    数量关系&Quantitative Relations&数学运算&Mathematical Calculations&不定方程问题&Indeterminate Equation Problems&&\\
    \midrule
    数量关系&Quantitative Relations&数学运算&Mathematical Calculations&不定方程问题&Indeterminate Equation Problems&普通不定方程&Ordinary Indeterminate Equations\\
    \midrule
    数量关系&Quantitative Relations&数学运算&Mathematical Calculations&不定方程问题&Indeterminate Equation Problems&不定方程组&Indeterminate Equation Groups\\
    \midrule
    数量关系&Quantitative Relations&数学运算&Mathematical Calculations&统筹规划问题&Coordination Planning Problems&&\\
    \midrule
    数量关系&Quantitative Relations&数学运算&Mathematical Calculations&数学运算-其他&Other&&\\
    \midrule
    数量关系&Quantitative Relations&数学运算&Mathematical Calculations&公倍数与公约数问题&Lcm And Gcd Problems&&\\
    \bottomrule
\end{tabular}
\end{CJK*}
}
\vfill\null

\subsection{Judgment\&Inference}
\resizebox{\columnwidth}{!}{%
\begin{CJK*}{UTF8}{gbsn}
\begin{tabular}{|c|c|c|c|c|c|c|c|}
    \toprule
    Level-1 Chinese&Level-1 English&Level-2 Chinese&Level-2 English&Level-3 Chinese&Level-3 English&Level-4 Chinese&Level-4 English \\
    \midrule
    判断推理&Judgment\&Inference&&&&&&\\
    \midrule
    判断推理&Judgment\&Inference&定义判断&Definitions And Judgements&&&&\\
    \midrule
    判断推理&Judgment\&Inference&定义判断&Definitions And Judgements&单定义&Single Definitions&&\\
    \midrule
    判断推理&Judgment\&Inference&定义判断&Definitions And Judgements&单定义&Single Definitions&主客体&Subject And Object\\
    \midrule
    判断推理&Judgment\&Inference&定义判断&Definitions And Judgements&单定义&Single Definitions&大前提&Major Premise\\
    \midrule
    判断推理&Judgment\&Inference&定义判断&Definitions And Judgements&单定义&Single Definitions&方式目的&Means And Purposes\\
    \midrule
    判断推理&Judgment\&Inference&定义判断&Definitions And Judgements&单定义&Single Definitions&原因结果&Causes And Results\\
    \midrule
    判断推理&Judgment\&Inference&定义判断&Definitions And Judgements&单定义&Single Definitions&单定义-其他句式&Other Patterns\\
    \midrule
    判断推理&Judgment\&Inference&定义判断&Definitions And Judgements&单定义&Single Definitions&故事类&Story\\
    \midrule
    判断推理&Judgment\&Inference&定义判断&Definitions And Judgements&单定义&Single Definitions&拆词&Word Parsing\\
    \midrule
    判断推理&Judgment\&Inference&定义判断&Definitions And Judgements&多定义&Multiple Definitions&&\\
    \midrule
    判断推理&Judgment\&Inference&定义判断&Definitions And Judgements&多定义&Multiple Definitions&常规问法&Conventional Question Forms\\
    \midrule
    判断推理&Judgment\&Inference&逻辑判断&Logical Judgements&&&&\\
    \midrule
    判断推理&Judgment\&Inference&逻辑判断&Logical Judgements&加强题型&Strengthening Question Types&&\\
    \midrule
    判断推理&Judgment\&Inference&逻辑判断&Logical Judgements&加强题型&Strengthening Question Types&搭桥&Building A Bridge\\
    \midrule
    判断推理&Judgment\&Inference&逻辑判断&Logical Judgements&加强题型&Strengthening Question Types&必要条件&Necessary Conditions\\
    \midrule
    判断推理&Judgment\&Inference&逻辑判断&Logical Judgements&加强题型&Strengthening Question Types&补充论据&Supplementing Evidence\\
    \midrule
    判断推理&Judgment\&Inference&逻辑判断&Logical Judgements&加强题型&Strengthening Question Types&加强选非题&Strengthening Wrong Choice Questions\\
    \midrule
    判断推理&Judgment\&Inference&逻辑判断&Logical Judgements&加强题型&Strengthening Question Types&加强-其他&Other\\
    \midrule
    判断推理&Judgment\&Inference&逻辑判断&Logical Judgements&削弱题型&Weakening Question Types&&\\
    \midrule
    判断推理&Judgment\&Inference&逻辑判断&Logical Judgements&削弱题型&Weakening Question Types&削弱论点&Weakening Viewpoints\\
    \midrule
    判断推理&Judgment\&Inference&逻辑判断&Logical Judgements&削弱题型&Weakening Question Types&拆桥&Dismantling The Bridge\\
    \midrule
    判断推理&Judgment\&Inference&逻辑判断&Logical Judgements&削弱题型&Weakening Question Types&他因削弱&Other Factors Weakening\\
    \midrule
    判断推理&Judgment\&Inference&逻辑判断&Logical Judgements&削弱题型&Weakening Question Types&削弱选非题&Weakening Wrong Choice Questions\\
    \midrule
    判断推理&Judgment\&Inference&逻辑判断&Logical Judgements&削弱题型&Weakening Question Types&削弱论据&Weakening Evidence\\
    \midrule
    判断推理&Judgment\&Inference&逻辑判断&Logical Judgements&削弱题型&Weakening Question Types&因果倒置&Reversing Cause And Effect\\
    \midrule
    判断推理&Judgment\&Inference&逻辑判断&Logical Judgements&削弱题型&Weakening Question Types&削弱-其他&Other\\
    \midrule
    判断推理&Judgment\&Inference&逻辑判断&Logical Judgements&翻译推理&Translation Reasoning&&\\
    \midrule
    判断推理&Judgment\&Inference&逻辑判断&Logical Judgements&翻译推理&Translation Reasoning&常规翻译&Conventional Translation\\
    \midrule
    判断推理&Judgment\&Inference&逻辑判断&Logical Judgements&翻译推理&Translation Reasoning&集合推理&Set Reasoning\\
    \midrule
    判断推理&Judgment\&Inference&逻辑判断&Logical Judgements&翻译推理&Translation Reasoning&推理形式&Reasoning Patterns\\
    \midrule
    判断推理&Judgment\&Inference&逻辑判断&Logical Judgements&翻译推理&Translation Reasoning&翻译推理-其他&Other\\
    \midrule
    判断推理&Judgment\&Inference&逻辑判断&Logical Judgements&组合排列-材料&Materials&&\\
    \midrule
    判断推理&Judgment\&Inference&逻辑判断&Logical Judgements&原因解释&Reason Explanations&&\\
    \midrule
    判断推理&Judgment\&Inference&类比推理&Analogical Reasoning&&&&\\
    \midrule
    判断推理&Judgment\&Inference&类比推理&Analogical Reasoning&语义关系&Semantic Relationships&&\\
    \midrule
    判断推理&Judgment\&Inference&类比推理&Analogical Reasoning&语义关系&Semantic Relationships&语义关系-近义关系&Synonyms\\
    \midrule
    判断推理&Judgment\&Inference&类比推理&Analogical Reasoning&语义关系&Semantic Relationships&语义关系-反义关系&Antonyms\\
    \midrule
    判断推理&Judgment\&Inference&类比推理&Analogical Reasoning&语义关系&Semantic Relationships&语义-其他&Other\\
    \midrule
    判断推理&Judgment\&Inference&类比推理&Analogical Reasoning&逻辑关系&Logical Relationships&&\\
    \midrule
    判断推理&Judgment\&Inference&类比推理&Analogical Reasoning&逻辑关系&Logical Relationships&逻辑关系-全同关系&Identical\\
    \midrule
    判断推理&Judgment\&Inference&类比推理&Analogical Reasoning&逻辑关系&Logical Relationships&逻辑关系-并列关系&Parallel\\
    \midrule
    判断推理&Judgment\&Inference&类比推理&Analogical Reasoning&逻辑关系&Logical Relationships&逻辑关系-交叉关系&Intersection\\
    \midrule
    判断推理&Judgment\&Inference&类比推理&Analogical Reasoning&逻辑关系&Logical Relationships&逻辑关系-包容关系&Inclusion\\
    \midrule
    判断推理&Judgment\&Inference&类比推理&Analogical Reasoning&逻辑关系&Logical Relationships&逻辑关系-对应关系&Correspondence\\
    \midrule
    判断推理&Judgment\&Inference&类比推理&Analogical Reasoning&拆分思维&Split Thinking&&\\
    \bottomrule
\end{tabular}
\end{CJK*}
}
\vfill\null

\subsection{Material Analysis}
\resizebox{\columnwidth}{!}{%
\begin{CJK*}{UTF8}{gbsn}
\begin{tabular}{|c|c|c|c|c|c|c|c|}
    \toprule
    Level-1 Chinese&Level-1 English&Level-2 Chinese&Level-2 English&Level-3 Chinese&Level-3 English \\
    \midrule
    资料分析&Material Analysis&&&&\\
    \midrule
    资料分析&Material Analysis&文字资料&Text Materials&&\\
    \midrule
    资料分析&Material Analysis&综合资料&Comprehensive Materials&&\\
    \midrule
    资料分析&Material Analysis&简单计算&Simple Calculations&&\\
    \midrule
    资料分析&Material Analysis&简单计算&Simple Calculations&直接找数&Directly Finding Numbers\\
    \midrule
    资料分析&Material Analysis&简单计算&Simple Calculations&简单加减计算&Simple Addition And Subtraction Calculations\\
    \midrule
    资料分析&Material Analysis&简单计算&Simple Calculations&排序类&Ordering\\
    \midrule
    资料分析&Material Analysis&基期与现期&Base And Present Periods&&\\
    \midrule
    资料分析&Material Analysis&基期与现期&Base And Present Periods&基期计算&Base Period Calculations\\
    \midrule
    资料分析&Material Analysis&基期与现期&Base And Present Periods&现期计算&Present Period Calculations\\
    \midrule
    资料分析&Material Analysis&基期与现期&Base And Present Periods&基期比较&Base Period Comparisons\\
    \midrule
    资料分析&Material Analysis&基期与现期&Base And Present Periods&间隔基期&Interval Base Periods\\
    \midrule
    资料分析&Material Analysis&基期与现期&Base And Present Periods&基期和差&Base Period Differences\\
    \midrule
    资料分析&Material Analysis&基期与现期&Base And Present Periods&现期追赶&Present Period Catching Up\\
    \midrule
    资料分析&Material Analysis&增长率&Growth Rates&&\\
    \midrule
    资料分析&Material Analysis&增长率&Growth Rates&一般增长率&General Growth Rate\\
    \midrule
    资料分析&Material Analysis&增长率&Growth Rates&混合增长率&Compound Growth Rate\\
    \midrule
    资料分析&Material Analysis&增长率&Growth Rates&间隔增长率&Interval Growth Rate\\
    \midrule
    资料分析&Material Analysis&增长率&Growth Rates&年均增长率&Annual Growth Rate\\
    \midrule
    资料分析&Material Analysis&增长量&Growth Amounts&&\\
    \midrule
    资料分析&Material Analysis&增长量&Growth Amounts&增长量计算&Growth Amount Calculations\\
    \midrule
    资料分析&Material Analysis&增长量&Growth Amounts&增长量比较&Growth Amount Comparisons\\
    \midrule
    资料分析&Material Analysis&增长量&Growth Amounts&间隔增长量&Interval Growth Amounts\\
    \midrule
    资料分析&Material Analysis&增长量&Growth Amounts&年均增长量&Annual Growth Amounts\\
    \midrule
    资料分析&Material Analysis&比重问题&Percentage Problems&&\\
    \midrule
    资料分析&Material Analysis&比重问题&Percentage Problems&现期比重&Present Period Percentage\\
    \midrule
    资料分析&Material Analysis&比重问题&Percentage Problems&基期比重&Base Period Percentage\\
    \midrule
    资料分析&Material Analysis&比重问题&Percentage Problems&两期比重&Two Period Percentages\\
    \midrule
    资料分析&Material Analysis&比重问题&Percentage Problems&混合比重&Mixed Percentages\\
    \midrule
    资料分析&Material Analysis&平均数问题&Average Problems&&\\
    \midrule
    资料分析&Material Analysis&平均数问题&Average Problems&基期平均数&Base Period Average\\
    \midrule
    资料分析&Material Analysis&平均数问题&Average Problems&现期平均数&Present Period Average\\
    \midrule
    资料分析&Material Analysis&平均数问题&Average Problems&平均数的增长率&Growth Rate Of Averages\\
    \midrule
    资料分析&Material Analysis&平均数问题&Average Problems&平均数的增长量&Growth Amount Of Averages\\
    \midrule
    资料分析&Material Analysis&平均数问题&Average Problems&两期平均数比较&Comparing Two Period Averages\\
    \midrule
    资料分析&Material Analysis&倍数与比值相关&Multiples And Ratios&&\\
    \midrule
    资料分析&Material Analysis&倍数与比值相关&Multiples And Ratios&基期倍数&Base Period Multiples\\
    \midrule
    资料分析&Material Analysis&倍数与比值相关&Multiples And Ratios&现期倍数&Present Period Multiples\\
    \midrule
    资料分析&Material Analysis&倍数与比值相关&Multiples And Ratios&比值计算&Ratio Calculations\\
    \midrule
    资料分析&Material Analysis&倍数与比值相关&Multiples And Ratios&比值比较&Ratio Comparisons\\
    \midrule
    资料分析&Material Analysis&综合分析&Comprehensive Analysis&&\\
    \bottomrule
\end{tabular}
\end{CJK*}
}
\vfill\null

\subsection{General Knowledge}
\resizebox{\columnwidth}{!}{%
\begin{CJK*}{UTF8}{gbsn}
\begin{tabular}{|c|c|c|c|c|c|c|c|}
    \toprule
    Level-1 Chinese&Level-1 English&Level-2 Chinese&Level-2 English&Level-3 Chinese&Level-3 English \\
    \midrule
    常识判断&General Knowledge&&&&\\
    \midrule
    常识判断&General Knowledge&政治常识&Political Knowledge&&\\
    \midrule
    常识判断&General Knowledge&政治常识&Political Knowledge&时政&Current Affairs\\
    \midrule
    常识判断&General Knowledge&政治常识&Political Knowledge&中国特色社会主义建设&Socialist Construction With Chinese Characteristics\\
    \midrule
    常识判断&General Knowledge&经济常识&Economic Knowledge&&\\
    \midrule
    常识判断&General Knowledge&经济常识&Economic Knowledge&宏观经济与调控政策&Macroeconomics And Regulation\\
    \midrule
    常识判断&General Knowledge&科技常识&Scientific Knowledge&&\\
    \midrule
    常识判断&General Knowledge&科技常识&Scientific Knowledge&物理常识&Physics Basics\\
    \midrule
    常识判断&General Knowledge&科技常识&Scientific Knowledge&化学常识&Chemistry Basics\\
    \midrule
    常识判断&General Knowledge&科技常识&Scientific Knowledge&生物常识&Biology Basics\\
    \midrule
    常识判断&General Knowledge&科技常识&Scientific Knowledge&科技理论与成就&Science And Technology Theories And Achievements\\
    \midrule
    常识判断&General Knowledge&科技常识&Scientific Knowledge&生活常识&Common Knowledge\\
    \midrule
    常识判断&General Knowledge&人文常识&Humanities Knowledge&&\\
    \midrule
    常识判断&General Knowledge&人文常识&Humanities Knowledge&中国历史&Chinese History\\
    \midrule
    常识判断&General Knowledge&人文常识&Humanities Knowledge&世界历史&World History\\
    \midrule
    常识判断&General Knowledge&人文常识&Humanities Knowledge&文学常识&Literature Basics\\
    \midrule
    常识判断&General Knowledge&人文常识&Humanities Knowledge&文化常识&Cultural Basics\\
    \midrule
    常识判断&General Knowledge&地理国情&Geography And National Conditions&&\\
    \midrule
    常识判断&General Knowledge&地理国情&Geography And National Conditions&自然常识&Natural Science Basics\\
    \midrule
    常识判断&General Knowledge&地理国情&Geography And National Conditions&国情社情&National Conditions\\
    \midrule
    常识判断&General Knowledge&法律常识&Legal Knowledge&&\\
    \midrule
    常识判断&General Knowledge&法律常识&Legal Knowledge&宪法&Constitution\\
    \midrule
    常识判断&General Knowledge&法律常识&Legal Knowledge&行政法&Administrative Law\\
    \midrule
    常识判断&General Knowledge&法律常识&Legal Knowledge&民法&Civil Law\\
    \midrule
    常识判断&General Knowledge&法律常识&Legal Knowledge&刑法&Criminal Law\\
    \midrule
    常识判断&General Knowledge&法律常识&Legal Knowledge&劳动法&Labor Law\\
    \midrule
    常识判断&General Knowledge&法律常识&Legal Knowledge&其他法律法规&Other Laws And Regulations\\
    \midrule
    常识判断&General Knowledge&法律常识&Legal Knowledge&民事诉讼法&Civil Procedure Law\\
    \midrule
    常识判断&General Knowledge&法律常识&Legal Knowledge&经济法&Economic Law\\
    \bottomrule
\end{tabular}
\end{CJK*}
}
\vfill\null

\end{document}